\documentclass[11pt]{article}
\usepackage{coling2018}
\usepackage[utf8]{inputenc}
\usepackage[T1]{fontenc}
\usepackage[english]{babel}
\usepackage{times}
\usepackage{url}
\usepackage{latexsym}
\usepackage{graphicx}
\usepackage{chronology}
\usepackage{wrapfig}
\usepackage[colorinlistoftodos]{todonotes}

\title{Diachronic word embeddings and semantic shifts: a survey}

\usepackage{latexsym}

\author{Andrey Kutuzov \hspace{10px} Lilja Øvrelid \hspace{10px} Terrence Szymanski\textsuperscript{$\diamondsuit$} \hspace{10px}  Erik Velldal  \\
[0.5ex] 
        University of Oslo, Norway\\
        {\tt\{andreku$\,$|$\,$liljao$\,$|$\,$erikve\}@ifi.uio.no}\\[0.5ex]
        \textsuperscript{$\diamondsuit$}ANZ, Melbourne, Australia \\
        {\tt terry.szymanski@gmail.com} 
        }  
  
\date{}

\begin{document}
\maketitle
\begin{abstract}
Recent years have witnessed a surge of publications aimed at tracing temporal changes in lexical semantics using distributional methods, particularly prediction-based word embedding models. 
However, this vein of research lacks the cohesion, common terminology and shared practices of more established areas of natural language processing. 
In this paper, we survey the current state of academic research related to diachronic word embeddings and semantic shifts detection. We start with discussing the notion of semantic shifts, and then continue with an overview of the existing methods for tracing such time-related shifts with word embedding models. We propose several axes along which these methods can be compared, and outline the main challenges before this emerging subfield of NLP, as well as prospects and possible applications.
\end{abstract}


\section{Introduction} \label{sec:intro}
%
\blfootnote{
    %
    %
    \hspace{-0.65cm}  
    This work is licensed under a Creative Commons 
    Attribution 4.0 International License.
    License details:
    \url{http://creativecommons.org/licenses/by/4.0/}
}

The meanings of words continuously change over time, reflecting complicated processes in language and society. 
Examples include both changes to the core meaning of words (like the word \textit{gay} shifting from meaning `carefree' to `homosexual' during the 20th century) and subtle shifts of cultural associations (like \textit{Iraq} or \textit{Syria} being associated with the concept of `war' after armed conflicts had started in these countries). Studying these types of changes in meaning enables researchers to learn more about human language and to extract temporal-dependent data from texts.

\begin{figure}
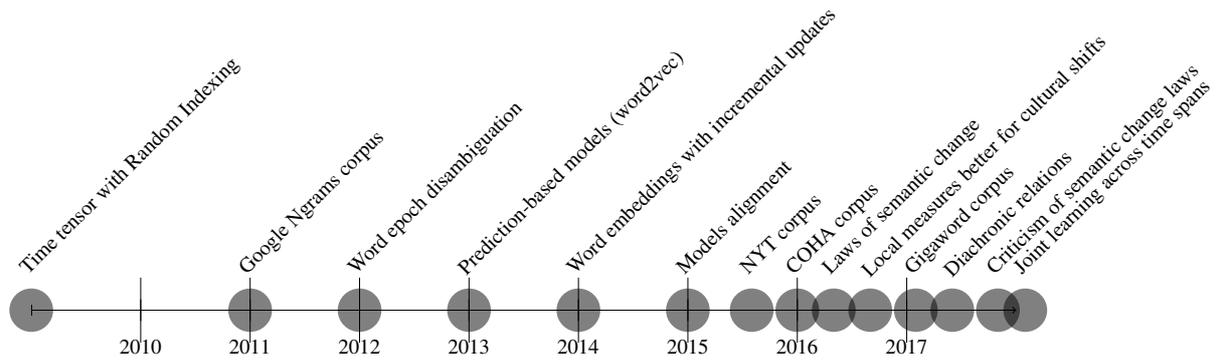

\begin{chronology}[1]{2009}{2017}{\textwidth}
\event{2009}{Time tensor with Random Indexing}
\event{2011}{Google Ngrams corpus}
\event{2012}{Word epoch disambiguation}
\event{2013}{Prediction-based models (word2vec)}
\event{2014}{Word embeddings with incremental updates}
\event{\decimaldate{01}{1}{2015}}{Models alignment}
\event{\decimaldate{01}{8}{2015}}{NYT corpus}
\event{\decimaldate{01}{1}{2016}}{COHA corpus}
\event{\decimaldate{01}{5}{2016}}{Laws of semantic change}
\event{\decimaldate{01}{9}{2016}}{Local measures better for cultural shifts}
\event{\decimaldate{01}{2}{2017}}{Gigaword corpus}
\event{\decimaldate{01}{6}{2017}}{Diachronic relations}
\event{\decimaldate{01}{11}{2017}}{Criticism of semantic change laws}
\event{\decimaldate{01}{02}{2018}}{Joint learning across time spans}
\end{chronology}
\caption{Distributional models in the task of tracing diachronic semantic shifts: research timeline}
\label{fig:timeline}
\end{figure}

The availability of large corpora and the development of computational semantics have given rise to a number of research initiatives trying to capture \textit{diachronic semantic shifts} in a data-driven way. Recently, \textit{word embeddings} \cite{Mikolov_representation:2013} have become a widely used input representation for this task. There are dozens of papers on the topic, mostly published after 2011 (we survey them in Section \ref{sec:automatic} and further below). However, this emerging field is highly heterogenous. There are at least three different research communities interested in it: natural language processing (and computational linguistics), information retrieval (and computer science in general), and political science. 
This is reflected in the terminology, which is far from being standardized.  One can find mentions of `temporal embeddings,' `diachronic embeddings,' `dynamic embeddings,' etc., depending on the background of a particular research group. 
The present survey paper attempts to describe this diversity, introduce some axes of comparison and outline main challenges which the practitioners face. Figure~\ref{fig:timeline} shows the timeline of events that influenced the research in this area: in the following sections we cover them in detail.

This survey is restricted in scope to research which traces semantic shifts using distributional word embedding models (that is, representing lexical meaning with dense vectors produced from co-occurrence data). We only briefly mention other data-driven approaches also employed to analyze temporal-labeled corpora (for example, topic modeling). Also, we do not cover syntactic shifts and other changes in the functions rather than meaning of words. 

The paper is structured as follows. In Section \ref{sec:shifts} we introduce the notion of `semantic shift' and provide some linguistic background for it. Section \ref{sec:automatic} aims to compare different approaches to the task of automatic detection of semantic shifts: in the choice of diachronic data, evaluation strategies, methodology of extracting semantic shifts from data, and the methods to compare word vectors across time spans. Sections \ref{sec:laws} and \ref{sec:relations} describe two particularly interesting results of diachronic embeddings research: namely, the statistical laws of semantic change and temporal semantic relations. In Section \ref{sec:applications} we outline possible applications of systems that trace semantic shifts. Section \ref{sec:challenges} presents open challenges which we believe to be most important for the field, and in Section \ref{sec:conclusion} we summarize and conclude.

\section{The concept of semantic shifts} \label{sec:shifts}
Human languages change over time, due to a variety of linguistic and non-linguistic factors and at all levels of linguistic analysis. In the field of theoretical (diachronic) linguistics, much attention has been devoted to expressing regularities of linguistic change. For instance, laws of phonological change have been formulated (e.g., Grimm's law or the great vowel shift) to account for changes in the linguistic sound system.  When it comes to lexical semantics, linguists have studied the evolution of word meaning over time, describing so-called lexical \textit{semantic shifts} or \textit{semantic change}, which \newcite{bloomfield} defines as ``innovations which change the lexical meaning rather than the grammatical function of a form.'' 

Historically, much of the theoretical work on semantic shifts has been devoted to documenting and categorizing various types of semantic shifts \cite{breal,stern1931meaning,bloomfield}.
The categorization found in \newcite{bloomfield} is arguably the most used and has inspired
a number of more recent studies \cite{blank1999historical,Geeraerts,traugott2001regularity}. \newcite{bloomfield} originally proposed nine classes of semantic shifts, six of which are complimentary pairs along a dimension. For instance, the pair `narrowing' -- `broadening' describes the observation that word meaning often changes to become either more specific or more general, e.g. Old English {\it mete} `food' becomes English {\it meat} `edible flesh,' or that the more general English word {\it dog} is derived from Middle English {\it dogge} which described a dog of a particular breed. \newcite{bloomfield} also describes change along the spectrum from positive to negative, describing the speaker's attitude as one of either degeneration or elevation, e.g. from Old English {\it cniht} 'boy, servant' to the more elevated {\it knight}.


The driving forces of semantic shifts are varied, but include linguistic, psychological, sociocultural or cultural/encyclopedic causes \cite{blank1999historical,grzega2007english}. Linguistic processes that cause semantic shifts generally involve the interaction between words of the vocabulary and their meanings. This may be illustrated by the process of ellipsis, whereby the meaning of one word is transferred to a word with which it frequently co-occurs, or by the need for discrimination of synonyms caused by lexical borrowings from other languages.  Semantic shifts may be also be caused by changes in the attitudes of speakers or in the general environment of the speakers.
Thus, semantic shifts are naturally separated into two important classes: linguistic drifts (slow and regular changes in core meaning of words) and cultural shifts (culturally determined changes in associations of a given word). Researchers studying semantic shifts from a computational point of view have shown the existence of this division empirically \cite{hamilton2016cultural}. In the traditional classification of \newcite{stern1931meaning}, the semantic shift category of \textit{substitution} describes a change that has a non-linguistic cause, namely that of technological progress. This may be exemplified by the word {\it car} which shifted its meaning from non-motorized vehicles after the introduction of the automobile. 

The availability of large corpora have enabled the development of new methodologies for the study of lexical semantic shifts within general linguistics \cite{traugottoxford}. A key assumption in much of this work is that changes in a word's collocational patterns reflect changes in word meaning \cite{Hilpert08}, thus providing a usage-based account of semantics \cite{Gries99}. For instance, \newcite{Kerremans} study the very recent neologism {\it detweet}, showing the development of two separate usages/meanings for this word (`to delete from twitter,' vs `to avoid tweeting') based on large amounts of web-crawled data. 
The usage-based view of lexical semantics aligns well with the assumptions underlying the distributional semantic approach \cite{firth1957synopsis} often employed in NLP . Here, the time spans studied are often considerably shorter (decades, rather than centuries) and we find that these  distributional methods seem well suited for monitoring the gradual process of meaning change. \newcite{baroni:2011}, for instance, showed that distributional models capture cultural shifts, like the word \textit{sleep} acquiring more negative connotations related to sleep disorders, when comparing its 1960s contexts to its 1990s contexts. 

To sum up, semantic shifts are often reflected in large corpora through change in the context of the word which is undergoing a shift, as measured by co-occurring words. It is thus natural to try to detect semantic shifts automatically, in a `data-driven' way. This vein of research is what we cover in the present survey. In the following sections, we overview the methods currently used for the automatic detection of semantic shifts and the recent academic achievements related to this problem.

\section{Tracing semantic shifts distributionally} \label{sec:automatic}
Conceptually, the task of  discovery of semantic shifts from data can be formulated as follows. Given corpora $[C_1, C_2, ...C_n]$ containing texts created in time periods $[1, 2,... n]$, the task is to locate words with different meaning in different time periods, or to locate the words which changed most. Other related tasks are possible: discovering general trends in semantic shifts (see Section \ref{sec:laws}) or tracing the dynamics of the relationships between words (see Section \ref{sec:relations}).  In the next subsections, we address several axes along which one can categorize the research on detecting semantic shifts with distributional models.  

\subsection{Sources of diachronic data for training and testing}
When automatically detecting semantic shifts, the types of generalizations we will be able to infer are influenced by properties of the textual data being used, such as the source of the datasets and the temporal granularity of the data. In this subsection we discuss the data choices made by researchers (of course, not pretending to cover the whole range of the diachronic corpora used).

\subsubsection{Training data}
The time unit (the granularity of the temporal dimension) can be chosen before slicing the text collection into subcorpora. Earlier works dealt mainly with long-term semantic shifts (spanning decades or even centuries), as they are easier to trace. One of the early examples is \newcite{sagi2011tracing} who studied differences between Early Middle, Late Middle and Early Modern English, using the Helsinki Corpus \cite{rissanen1993helsinki}.

The release of the Google Books Ngrams corpus\footnote{\url{https://books.google.com/ngrams}} played an important role in the development of the field and spurred work on the new discipline of `culturomics,' studying human culture through digital media \cite{Michel:2011}. \newcite{mihalcea:2012} used this dataset to detect differences in word usage and meaning across 50-years time spans, while \newcite{baroni:2011} compared word meanings in the 1960s and in the 1990s, achieving good correlation with human judgments. Unfortunately, Google Ngrams is inherently limited in that it does not contain full texts. However, for many cases, this corpus was enough, and its usage as the source of diachronic data continued in \newcite{biemann2014} (employing syntactic ngrams), who detected word sense changes over several different time periods spanning from 3 to 200 years.  

In more recent work, time spans tend to decrease in size and become more granular. In general, corpora with smaller time spans are useful for analyzing socio-cultural semantic shifts, while corpora with longer spans are necessary for the study of linguistically motivated semantic shifts. As researchers are attempting to trace increasingly subtle cultural semantic shifts (more relevant for practical tasks), the granularity of time spans is decreasing: for example, \newcite{kim2014temporal} and \newcite{cheng:2016} analyzed the \textit{yearly} changes of words. Note that, instead of using granular `bins', time can also be represented as a continuous differentiable value \cite{rosenfeld:2018}.

In addition to the Google Ngrams dataset (with granularity of 5 years), \newcite{kulkarni2015statistically} used Amazon Movie Reviews (with granularity of 1 year) and Twitter data (with granularity of 1 month). Their results indicated that computational methods for the detection of semantic shifts  can be robustly applied to time spans less than a decade. \newcite{zhang2015} used another yearly text collection, the New-York Times Annotated Corpus \cite{sandhaus2008new}, again managing to trace subtle semantic shifts. The same corpus was employed by \newcite{szymanski:17}, with 21 separate models, one for each year from 1987 to 2007, and to some extent by \newcite{Yao:2018}, who crawled the NYT web site to get 27 yearly subcorpora (from 1990 to 2016). The inventory of diachronic corpora used in tracing semantic shifts was expanded by \newcite{eger2016}, who used the Corpus of Historical American (COHA\footnote{\url{http://corpus.byu.edu/coha/}}), with time slices equal to one decade. \newcite{jurafsky2016} continued the usage of COHA (along with the Google Ngrams corpus). \newcite{kutuzov2017a} started to employ the yearly slices of the English Gigaword corpus \cite{Gigaword:11} in the analysis of cultural semantic drift related to armed conflicts. 



\subsubsection{Test sets}
Diachronic corpora are needed not only as a source of \textit{training} data for developing semantic shift detection systems, but also as a source of \textit{test} sets to evaluate such systems. In this case, however, the situation is more complicated. Ideally, diachronic approaches should be evaluated on human-annotated lists of semantically shifted words (ranked by the degree of the shift). However, such gold standard data is difficult to obtain, even for English, let alone for other languages. General linguistics research on language change like that of  \newcite{traugott2001regularity} and others usually contain only a small number of hand-picked examples, which is not sufficient to properly evaluate an automatic unsupervised system.

Various ways of overcoming this problem have been proposed. For example, \newcite{mihalcea:2012} evaluated the ability of a system to detect the time span that specific contexts of a word undergoing a shift belong to (\textit{word epoch disambiguation}). A similar problem was offered as SemEval-2015 Task 7: `Diachronic Text Evaluation' \cite{popescu:2015}. Another possible evaluation method is so-called cross-time alignment, where a system has to find equivalents for certain words in different time periods (for example, `Obama' in 2015 corresponds to `Trump' in 2017). There exist several datasets containing such temporal equivalents for English \cite{Yao:2018}. Yet another evaluation strategy is to use the detected diachronic semantic shifts to trace or predict real-world events like armed conflicts \cite{kutuzov2017a}. Unfortunately, all these evaluation methods still require the existence of large manually annotated semantic shift datasets. The work to properly create and curate such datasets is in its infancy. 

One reported approach to avoid this requirement is borrowed from research on word sense disambiguation and consists of making a synthetic task by merging two real words together and then modifying the training and test data according to a predefined sense-shifting function. \newcite{rosenfeld:2018} successfully employed this approach to evaluate their system; however, it still operates on synthetic words, limiting the ability of this evaluation scheme to measure the models' performance with regards to real semantic shift data. Thus, the problem of evaluating semantic shift detection approaches is far from being solved, and practitioners often rely on self-created test sets, or even simply manually inspecting the results.

\subsection{Methodology of extracting semantic shifts from data}

After settling on a diachronic data set to be used in the system, one has to choose the methods to analyze it. Before the broad adoption of word embedding models, it was quite common to use change in raw word frequencies in order to trace semantic shifts or other kinds of linguistic change; see, among others, \newcite{juola:2003}, \newcite{hilpert:2009}, \newcite{Michel:2011}, \newcite{lijffijt:2012}, or \newcite{choi2012predicting} for frequency analysis of words in web search queries. Researchers also studied the increase or decrease in the frequency of a word $A$ collocating with another word $B$ over time, and based on this inferred changes in the meaning of $A$ \cite{heyer2009change}. 

However, it is clear that semantic shifts are not always accompanied with changes in word frequency (or this connection may be very subtle and non-direct). Thus, if one were able to more directly model word meaning, such an approach should be superior to frequency-proxied methods. A number of recent publications have showed that \textit{distributional word representations} \cite{turney2010frequency,baroni2014don} provide an efficient way to solve these tasks. They represent meaning with sparse or dense (embedding) vectors, produced from word co-occurrence counts. Although conceptually the source of the data for these models is still word frequencies, they `compress' this information into continuous lexical representations which are both efficient and convenient to work with. 
Indeed, \newcite{kulkarni2015statistically} explicitly demonstrated that distributional models outperform the frequency-based methods in detecting semantic shifts. They managed to trace semantic shifts more precisely and with greater explanatory power. One of the examples from their work is the semantic evolution of the word \textit{gay}: through time, its nearest semantic neighbors changed, manifesting the gradual move away from the sense of `cheerful' to the sense of `homosexual.' 


In fact, distributional models were being used in diachronic research long before the paper of \newcite{kulkarni2015statistically}, although there was no rigorous comparison to the frequentist methods. Already in 2009, it was proposed that one can use distributional methods to detect semantic shifts in a quantitative way. The pioneering work by \newcite{jurgens_event} described an insightful conceptualization of a sequence of distributional model updates through time: it is effectively a Word:Semantic Vector:Time tensor, in the sense that each word in a distributional model possesses a set of semantic vectors for each time span we are interested in. 
It paved the way for quantitatively comparing not only words with regard to their meaning, but also different stages in the development of word meaning over time. 

\newcite{jurgens_event} employed the \textit{Random Indexing} (RI) algorithm \cite{kanerva2000random} to create word vectors.  Two years later, \newcite{baroni:2011} used explicit count-based models, consisting of sparse co-occurrence matrices weighted by Local Mutual Information, while \newcite{sagi2011tracing} turned to Latent Semantic Analysis \cite{deerwester1990indexing}. In \newcite{basile2014analysing}, an extension to RI dubbed \textit{Temporal Random Indexing} (TRI) was proposed. However, no quantitative evaluation of this approach was offered (only a few hand-picked examples based on the Italian texts from the \textit{Gutenberg Project}), and thus it is unclear whether TRI is any better than other distributional models for the task of semantic shift  detection.

Further on, the diversity of the employed methods started to increase. For example, \newcite{biemann2014} analyzed clusters of the word similarity graph in the subcorpora corresponding to different time periods. Their distributional model consisted of lexical nodes in the graphs connected with weighted edges. The weights corresponded to the number of shared most salient syntactic dependency contexts, where saliency was determined by co-occurrence counts scaled by Mutual Information (MI). Importantly, they were able to detect not only the mere fact of a semantic shift, but also its type: the birth of a new sense, splitting of an old sense into several new ones, or merging of several senses into one. Thus, this work goes into a much less represented class of `fine-grained' approaches to semantic shift detection. It is also important that \newcite{biemann2014} handle natively the issue of polysemous words, putting the much-neglected problem of word senses in the spotlight.

The work of \newcite{kim2014temporal} was seminal in the sense that it is arguably the first one employing \textit{prediction-based word embedding models} to trace diachronic semantic shifts. Particularly, they used incremental updates (see below) and Continuous Skipgram with negative sampling (SGNS) \cite{mikolov2013efficient}.\footnote{Continuous Bag-of-Words (CBOW) from the same paper is another popular choice for learning semantic vectors.} 
\newcite{jurafsky2016} showed the superiority of SGNS over explicit PPMI-based distributional models in semantic shifts analysis, although they noted that low-rank SVD approximations \cite{bullinaria2007extracting} can perform on par with SGNS, especially on smaller datasets. Since then, the majority of publications in the field started using dense word representations: either in the form of SVD-factorized PPMI matrices, or in the form of prediction-based shallow neural models like SGNS.\footnote{\newcite{levy2014neural} showed that these two approaches are  equivalent from the mathematical point of view.}

There are some works employing other distributional approaches to semantic shifts detection. For instance, there is a strong vein of research based on dynamic topic modeling \cite{blei2006dynamic,Wang:2006}, which learns the evolution of topics over time. 
In \newcite{wijaya2011understanding}, it helped solve a typical digital humanities task of finding traces of real-world events in the texts. \newcite{heyer2016modeling} employed topic analysis to trace the so-called `context volatility' of words. In the political science, topic models are also sometimes used as proxies to social trends developing over time: for example, \newcite{mueller_rauh_2017} employed LDA to predict timing of civil wars and armed conflicts. \newcite{frermann:2016} drew on these ideas to trace diachronic word senses development. But most scholars nowadays seem to prefer parametric distributional models, particularly prediction-based embedding algorithms like SGNS, CBOW or GloVe \cite{pennington2014glove}. Following their widespread adoption in NLP in general, they have become the dominant representations for the analysis of diachronic semantic shifts as well. 

\subsection{Comparing vectors across time} \label{subsec:alignment}
It is rather straightforward to train separate word embedding models using time-specific corpora containing texts from several different time periods. As a consequence, these models are also time-specific. However, it is not that straightforward to compare word vectors across different models.

It usually does not make sense to, for example, directly calculate cosine similarities between embeddings of one and the same word in two different models. The reason is that most modern word embedding algorithms are inherently stochastic and the resulting embedding sets are invariant under rotation. Thus, even when trained on the same data, separate learning runs will produce entirely different numerical vectors (though with roughly the same pairwise similarities between vectors for particular words). This is expressed even stronger for models trained on different corpora. It means that even if word meaning is completely stable, the direct cosine similarity between its vectors from different time periods can still be quite low, simply because the random initializations of the two models were different. 
To alleviate this, \newcite{kulkarni2015statistically} suggested that before calculating similarities, one should first \textit{align} the models to fit them in one vector space, using linear transformations preserving general vector space structure. After that, cosine similarities across models become meaningful and can be used as indicators of semantic shifts. They also proposed constructing the time series of a word embedding over time, which allows for the detection of `bursts' in its meaning with the \textit{Mean Shift} model \cite{taylor2000change}. Notably, almost simultaneously the idea of aligning diachronic word embedding models using a distance-preserving projection technique was proposed by \newcite{zhang2015}. Later, \newcite{zhang:2016} expanded on this by adding the so called `local anchors': that is, they used both linear projections for the whole models and small sets of nearest neighbors for mapping the query words to their correct temporal counterparts.

Instead of aligning their diachronic models using linear transformations, \newcite{eger2016} compared word meaning using so-called `second-order embeddings,' that is, the vectors of words' similarities to all other words in the shared vocabulary of all models. This approach does not require any transformations: basically, one simply analyzes the word's position compared to other words. At the same time, \newcite{jurafsky2016} and \newcite{hamilton2016cultural} showed that these two approaches can be used simultaneously:  they employed both `second order embeddings' and orthogonal Procrustes transformations to align diachronic models. 

Recently, it was shown in \newcite{bamler2017dynamic} (`\textit{dynamic skip-gram}' model) and \newcite{Yao:2018} (`\textit{dynamic Word2Vec}' model) that it is possible to learn the word embeddings across several time periods jointly, enforcing alignment across all of them simultaneously, and positioning all the models in the same vector space in one step. This develops the idea of model alignment even further and eliminates the need to first learn separate embeddings for each time period, and then align subsequent model pairs. \newcite{bamler2017dynamic} additionally describe two variations of their approach: a) for the cases when data slices arrive sequentially, as in streaming applications, where one can not use future observations, and b) for the cases when data slices are available all at once, allowing for training on the whole sequence from the very beginning.  A similar approach is taken by \newcite{rosenfeld:2018} who train a deep neural network on word and time representations. Word vectors in this setup turn into linear transformations applied to a continuous time variable, and thus producing an embedding of word $w$ at time $t$.

Yet another way to make the models comparable is made possible by the fact that prediction-based word embedding approaches (as well as RI) allow for incremental updates of the models with new data without any modifications. This is not the case for the traditional explicit count-based algorithms, which usually require a computationally expensive dimensionality reduction step. \newcite{kim2014temporal} proposed the idea of \textit{incrementally updated diachronic embedding models}: that is, they train a model on the year $y_i$, and then the model for the  year $y_{i+1}$ is initialized with the word vectors from $y_i$. This can be considered as an alternative to model alignment: instead of aligning models trained from scratch on different time periods, one starts with training a model on the diachronically first period, and then updates this same model with the data from the successive time periods, saving its state each time. Thus, all the models are inherently related to each other, which, again, makes it possible to directly calculate cosine similarities between the same word in different time period models, or at least makes the models more comparable. 

Several works have appeared recently which aim to address the technical issues accompanying this approach of incremental updating. Among others, \newcite{peng2017incrementally} described a novel method of incrementally learning the \textit{hierarchical softmax} function for the CBOW and Continuous Skipgram algorithms. In this way, one can update word embedding models with new data and new vocabulary much more efficiently, achieving faster training than when doing it from scratch, while at the same time preserving comparable performance. Continuing this line of research, \newcite{kaji2017incremental} proposed a conceptually similar incremental extension for \textit{negative sampling}, which is a method of training examples selection, widely used with prediction-based models as a faster replacement for \textit{hierarchical softmax}.

Even after the models for different time periods are made comparable in this or that way, one still has to choose the exact method of comparing word vectors across these models. \newcite{jurafsky2016} and \newcite{hamilton2016cultural} made an important observation that the distinction between linguistic and cultural semantic shifts is correlated with the distinction between \textit{global} and \textit{local} embedding comparison methods. The former take into account the whole model (for example, `second-order embeddings,' when we compare the word's similarities to all other words in the lexicon), while the latter focus on the word's immediate neighborhood (for example, when comparing the lists of $k$ nearest neighbors). They concluded that global measures are sensitive to regular processes of linguistic shifts, while local measures are better suited to detect slight cultural shifts in word meaning. Thus, the choice of particular embedding comparison approach should depend on what type of semantic shifts one seeks to detect.

\section{Laws of semantic change} \label{sec:laws}
The use of diachronic word embeddings for studying the dynamics of word meaning has resulted in several hypothesized `laws' of semantic change. We review some of these law-like generalizations below, before finally describing a study that questions their validity. 

\newcite{dubossarsky:2015} experimented with K-means clustering applied to SGNS embeddings trained for  evenly sized yearly samples for the period 1850--2009. They found that the degree of semantic change for a given word -- quantified as the change in self-similarity over time -- negatively correlates with its distance to the centroid of its cluster. They proposed that the likelihood for semantic shift correlates with the degree of prototypicality (the \textit{`law of prototypicality'} in \newcite{dubossarsky:2017}).

Another relevant study is reported by \newcite{eger2016}, based on two different graph models; one being a time-series model relating embeddings across time periods to model semantic shifts and the other modeling the self-similarity of words across time. Experiments were performed with time-indexed historical corpora of English, German and Latin, using time-periods corresponding to decades, years and centuries, respectively. To enable comparison of embeddings across time, second-order embeddings encoding similarities to other words were used, as described in \ref{subsec:alignment}, limited to the `core vocabulary' (words occurring at least 100 times in all time periods). Based on linear relationships observed in the graphs, \newcite{eger2016} postulate two `laws' of semantic change:

\begin{enumerate}
\item word vectors can be expressed as linear combinations of their neighbors in previous time periods;
\item the meaning of words tend to decay linearly in time, in terms of the similarity of a word to itself; this is in line with the `\textit{law of differentiation}' proposed by \newcite{xu2015computational}.
\end{enumerate}

In another study, \newcite{jurafsky2016} considered historical corpora for English, German, French and Chinese, spanning 200 years and using time spans of decades. The goal was to investigate the role of frequency and polysemy with respect to semantic shifts. As in \newcite{eger2016}, the rate of semantic change was quantified by self-similarity across time-points (with words represented by Procrustes-aligned SVD embeddings). Through a regression analysis, \newcite{jurafsky2016} investigated how the change rates correlate with frequency and polysemy, and proposed another two `laws':

\begin{enumerate}
\item frequent words change more slowly (`\textit{the law of conformity}');
\item polysemous words (controlled for frequency) change more quickly (`\textit{the law of innovation}').
\end{enumerate}

\newcite{Azarbonyad:2017} showed that these laws (at least the law of conformity) hold not only for diachronic corpora, but also for other `viewpoints': for example, semantic shifts across models trained on texts produced by different political actors or written in different genres \cite{kutuzov_2016_registers}. However, the temporal dimension allows for a view of the corpora under analysis as a sequence, making the notion of `semantic shift' more meaningful. 

Later, \newcite{dubossarsky:2017} questioned the validity of some of these proposed `laws' of semantic change. In a series of replication and control experiments, they demonstrated that some of the regularities observed in previous studies are largely artifacts of the models used and frequency effects. In particular, they considered 10-year bins comprising equally sized yearly samples from Google Books 5-grams of English fiction for the period 1990--1999. For control experiments, they constructed two additional data sets; one with chronologically shuffled data where each bin contains data from all decades evenly distributed, and one synchronous variant containing repeated random samples from the year 1999 alone. Any measured semantic shifts within these two alternative data sets would have to be due to random sampling noise. 

\newcite{dubossarsky:2017} performed experiments using raw co-occurrence counts, PPMI weighted counts, and SVD transformations (Procrustes aligned), and conclude that the `laws' proposed in previous studies -- that semantic change is correlated with frequency, polysemy \cite{jurafsky2016} and prototypicality \cite{dubossarsky:2015} -- are not valid as they are also observed in the control conditions. \newcite{dubossarsky:2017} suggested that these spurious effects are instead due to the type of word representation used -- count vectors -- and that semantic shifts must be explained by a more diverse set of factors than distributional ones alone. Thus, the discussion on the existence of the `laws of semantic change' manifested by distributional trends is still open. 

\section{Diachronic semantic relations} \label{sec:relations}
Word embedding models are known to successfully capture complex \textit{relationships} between concepts, as manifested in the well-known word analogies task \cite{mikolov2013efficient}, where a model must `solve' equations of the form `A is to B is as C is to what?' A famous example is the distributional model capturing the fact that the relation between `\textit{man}' and `\textit{woman}' is the same as between `\textit{king}' and `\textit{queen}' (by adding and subtracting the corresponding word vectors). Thus, it is a natural development to investigate whether changes in semantic relationships across time can also be traced by looking at the diachronic development of distributional models.

\newcite{zhang2015} considered the \textit{temporal correspondences problem}, wherein the objective is to identify the word in a target time period which corresponds to a query term in the source time period (for example, given the query term \textit{iPod} in the 2000s, the counterpart term in the 1980s time period is \textit{Walkman}). This is proposed as a means to improve the results of information retrieval from document collections with significant time spans. \newcite{szymanski:17} frames this as the \textit{temporal word analogy} problem, extending the word analogies concept into the temporal dimension. This work shows that diachronic word embeddings can successfully model relations like `word $w_1$ at time period $t_\alpha$ is like word $w_2$ at time period $t_\beta$'. To this end, embedding models trained on different time periods are aligned using linear transformations. Then, the temporal analogies are solved by simply finding out which word vector in the time period $t_\beta$ is the closest to the vector of $w_1$ in the time period $t_\alpha$. 

A variation of this task was studied in \newcite{rosin2017learning}, where the authors learn the relatedness of words over time, 
answering queries like `in which time period were the words \textit{Obama} and \textit{president} maximally related'.  This technique can be used for a more efficient user query expansion in general-purpose search engines. 
\newcite{kutuzov2017b} modeled a different semantic relation: `words $w_1$ and $w_2$ at time period $t_\alpha$ are in the same semantic relation as words $w_3$ and $w_4$ at time period $t_\beta$'. To trace the temporal dynamics of these relations, they re-applied linear projections learned on sets of $w_1$ and $w_2$ pairs from the model for the period $t_n$ to the model trained on the subsequent time period $t_{n+1}$. This was used to solve the task of detecting lasting or emerging armed conflicts and the violent groups involved in these conflicts. 


\section{Applications} \label{sec:applications}
Applications of diachronic word embeddings approaches can generally be grouped into two broad categories: \textit{linguistic studies} which investigate the how and why of semantic shifts, and \textit{event detection} approaches which mine text data for actionable purposes.

The first category generally involves corpora with longer time spans, since linguistic changes happen at a relatively slow pace. Some examples falling into this category include tracking semantic drift of particular words \cite{kulkarni2015statistically} or of word sentiment \cite{hamilton2016sentiment}, identifying the breakpoints between epochs \cite{sagi2011tracing,mihalcea:2012}, studying the laws of semantic change at scale \cite{hamilton2016cultural} and finding different words with similar meanings at different points in time \cite{szymanski:17}. This has been held up as a good use case of deep learning for  research in computational linguistics \cite{manning2015}, and there are opportunities for future work applying diachronic word embeddings not only in the field of historical linguistics, but also in related areas like sociolinguistics and digital humanities.

The second category involves mining texts for cultural semantic shifts (usually on shorter time spans) indicating real-world events. Examples of this category are temporal information retrieval \cite{rosin2017learning}, predicting civil turmoils \cite{kutuzov2017a,mueller_rauh_2017}, or tracing the popularity of entities using norms of word vectors \cite{Yao:2018}. They can potentially be employed to improve user experience in production systems or for policy-making in governmental structures.

We believe that the near future will see a more diverse landscape of applications for diachronic word embeddings, especially related to the real-time analysis of large-scale news streams. `Between the lines,' these data sources contain a tremendous amount of information about processes in our world, manifested in semantic shifts of various sorts. The task of researchers is to reveal this information and make it reliable and practically useful.  

\section{Open challenges} \label{sec:challenges}
The study of temporal aspects of semantic shifts using distributional models (including word embeddings) is far from being a solved problem. The field still has a considerable number of open challenges. Below we briefly describe the most demanding ones.

\begin{itemize}
\item The existing methods should be expanded to a \textit{wider scope of languages}. \newcite{jurafsky2016}, \newcite{kutuzov2018two} and others have started to analyze other languages, but the overwhelming majority of publications still apply only to English corpora. It might be the case that the best methodologies are the same for different languages, but this should be shown empirically.
\item There is a clear need to devise algorithms that work on \textit{small datasets}, as they are very common in historical linguistics, digital humanities, and similar disciplines.
\item Carefully designed and robust \textit{gold standard test sets} of semantic shifts (of different kinds) should be created. This is a difficult task in itself, but the experience from synchronic word embeddings evaluation \cite{hill:2015} and other NLP areas proves that it is possible. 
\item There is a need for rigorous \textit{formal mathematical models of diachronic embeddings}. Arguably, this will follow the vein of research in joint learning across several time spans, started by \newcite{bamler2017dynamic} and \newcite{Yao:2018}, but other directions are also open.
\item Most current studies stop after stating the simple fact that a semantic shift has occurred. However, more detailed analysis of the nature of the shift is needed. This includes:
\begin{enumerate}
\item \textit{Sub-classification of types of semantic shifts} (broadening, narrowing, etc). This problem was to some degree  addressed by \newcite{biemann2014}, but much more work is certainly required to empirically test classification schemes proposed in much of the theoretical work described in Section \ref{sec:shifts}.
\item \textit{Identifying the source of a shift} (for example, linguistic or extra-linguistic causes). This causation detection is closely linked to the division between linguistic drifts and cultural shifts, as proposed in \newcite{hamilton2016cultural}.
\item \textit{Quantifying the weight of senses} acquired over time. Many words are polysemous, and the relative importance of senses is flexible \cite{frermann:2016}. The issue of handling senses is central for detecting semantic shifts, but most of the algorithms described in this survey are not sense-aware. To address this, methods from sense embeddings research \cite{bartunov2016breaking} might be employed.
\item \textit{Identifying groups of words that shift together} in correlated ways. Some work in this direction was started in \newcite{dubossarsky2016verbs}, who showed that verbs change more than nouns, and nouns change more than adjectives. This is also naturally related to proving the (non-)existence of the `laws of semantic change' (see Section \ref{sec:laws}).
\end{enumerate}
\item Last but not least, we believe that the community around diachronic word embeddings research severely lacks relevant forums, like \textit{topical workshops} or \textit{shared tasks}. Diachronic text evaluation tasks like the one at \textit{SemEval-2015} \cite{popescu:2015} are important but not enough, since they focus on identifying the time period when a text was authored, not the process of shifting meanings of a word. Organizing such events can promote the field and help address many of the challenges described above.
\end{itemize}

\section{Summary} \label{sec:conclusion}
We have presented an outline of the current research related to computational detection of semantic shifts using diachronic (temporal) word embeddings. We covered the linguistic nature of semantic shifts, the typical sources of diachronic data and the distributional approaches used to model it, from frequentist methods to contemporary prediction-based models. To sum up, Figure~\ref{fig:timeline} shows the timeline of events that have been influential in the development of research in this area: introducing concepts, usage of corpora and important findings.

This emerging field is still relatively new, and although recent years has seen a string of significant discoveries and academic interchange, much of the research still appears slightly fragmented, not least due to the lack of dedicated venues like workshops, special issues, or shared tasks. We hope that this survey will be useful to those who want to understand how this field has developed, and gain an overview of what defines the current state-of-the-art and what challenges lie ahead.

\section*{Acknowledgements}
We thank William Hamilton, Haim Dubossarsky and Chris Biemann for their helpful feedback during the preparation of this survey. All possible mistakes remain the sole responsibility of the authors.

\bibliographystyle{acl}
\bibliography{dia}

\end{document}